\documentclass[11pt]{article}
\usepackage[preprint]{acl}

\usepackage{times}
\usepackage{graphicx}
\usepackage{natbib}
\usepackage{caption}
\usepackage{algorithm}
\usepackage{algpseudocode}
\usepackage{amsmath}
\usepackage{tabularx}
\usepackage{arydshln}
\usepackage{booktabs}
\usepackage{multirow}
\usepackage{array}
\usepackage{amssymb}
\usepackage{newfloat}
\usepackage{listings}
\usepackage{xcolor}

\usepackage[T1]{fontenc}
\usepackage[utf8]{inputenc}
\usepackage{microtype}
\usepackage{inconsolata}

\DeclareCaptionStyle{ruled}{labelfont=normalfont,labelsep=colon,strut=off}
\floatstyle{ruled}
\newfloat{listing}{tb}{lst}{}
\floatname{listing}{Listing}

\title{Can a Small Model Learn to Look Before It Leaps? Dynamic Learning and Proactive Correction for Hallucination Detection}
\author{
  Zepeng Bao$^{1}$, 
  Shen Zhou$^{1}$, 
  Qiankun Pi$^{1}$, 
  Jianhao Chen$^{1,2}$, 
  Mayi Xu$^{1}$ \\
  \textbf{Ming Zhong}$^{1}$, 
  \textbf{Yuanyuan Zhu}$^{1}$, 
  \textbf{Tieyun Qian}$^{1,2}$\thanks{Corresponding author} \\
  $^{1}$School of Computer Science, Wuhan University, China \\
  $^{2}$Zhongguancun Academy, Beijing, China \\
  \texttt{\{zepengbao, qty\}@whu.edu.cn}
}

\usepackage{bibentry}

\begin{document}

\maketitle
\begin{abstract}
Hallucination in large language models (LLMs) remains a critical barrier to their safe deployment. 
For hallucination detection to be practical in real-world scenarios, the use of efficient small models is essential to ensure low latency and minimal resource consumption. However, existing methods rely on fixed verification strategies, where simply tuning small models to mimic fixed verification trajectories fails to capture the adaptability required for diverse hallucination patterns, thereby inducing planning instability.
To address this limitation, we propose a ``Learning to Evaluate and Adaptively Plan'' (LEAP) framework, which shifts hallucination detection from fixed execution to dynamic strategy learning. 
Specifically, LEAP first employs a powerful teacher model to iteratively explore and refine verification strategies through a failure-driven loop. This dynamic planning capability is then distilled into an efficient student model, augmented by a novel proactive correction mechanism that enables the model to evaluate and optimize its verification strategy before execution.
Experiments on three benchmarks demonstrate that LEAP outperforms state-of-the-art methods, offering an effective and scalable solution for reliable hallucination detection.
\end{abstract}
\section{Introduction}
Hallucination undermines the reliability of large language models (LLMs) through the generation of factually incorrect or fabricated content.
This issue poses severe risks in high-stakes domains such as medicine and law~\cite{ji2023survey, zhang2023siren}. Therefore, equipping LLMs with the ability to discern the veracity of their own outputs via \textbf{hallucination detection} has become a critical prerequisite for safe deployment~\cite{huang2025survey}.

\begin{figure}[t]
    \centering 
    \includegraphics[width=0.48\textwidth]{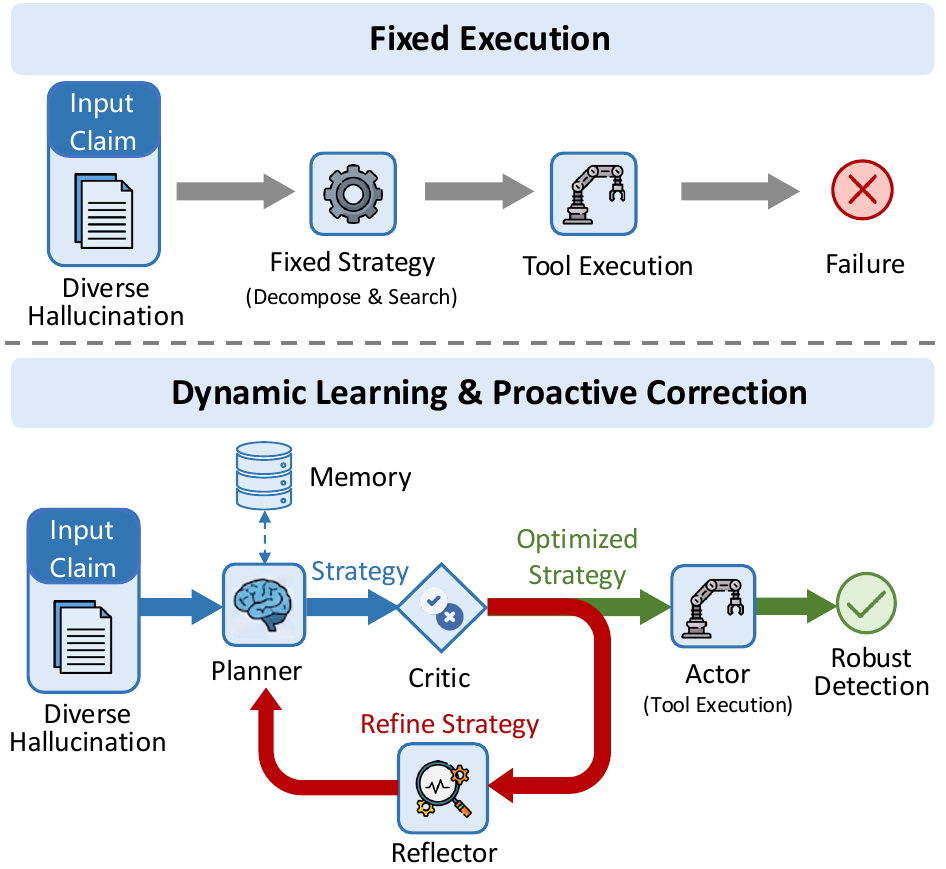} 
    \caption{Fixed strategies and dynamic strategies for hallucination detection on diverse claims.}
    \label{fig:intro}
\end{figure}

Existing detection methods typically fall into two paradigms: intrinsic self-check and tool-augmented verification. The former leverages models' internal signals like token probabilities \cite{ren2023ppl,Kuhn2023SemanticUncertainty} or patterns in activation states \cite{du2024haloscope, bazarova2025hallucination} for hallucination flagging. However, these methods may fail when the model is confidently wrong.

This constraint has spurred the development of tool-augmented methods \cite{chern2023factool, wei2024long}, which verify claims by retrieving external evidence. 
For such tool-augmented detection to be practical in real-world applications, there is a growing emphasis on using efficient small models \cite{belyi-etal-2025-luna}. Their low latency and minimal resource requirements make them ideal for real-time monitoring and on-device deployment. However, the limited parameter scale of these models introduces a significant performance bottleneck. 

To compensate for their restricted reasoning capabilities, existing frameworks typically resort to \textbf{predefined and fixed verification strategies}. As shown in Figure \ref{fig:intro}, these methods execute the same ``search-and-verify'' workflow, regardless of whether a claim involves simple facts or complex causal relationships. 
This fixity lacks the adaptability required for diverse hallucination patterns, often leading to inappropriate tool calls and detection failures.
Furthermore, even when specifically tuned~\cite{cheng2024small}, they also suffer from planning instability due to mimicking fixed trajectories over adaptive reasoning and may generate plausible but ineffective verification plans when facing complex claims, as shown in Appendix~\ref{sec:case_study}.

To bridge this gap, we argue that simply tuning small models to mimic fixed tool-use trajectories is insufficient, as it fails to capture the underlying reasoning logic required for diverse claims.
Instead, robust hallucination detection requires a paradigm shift from executing fixed process to planning adaptive dynamic strategies. However, implementing this shift presents two core challenges. First, since optimal verification paths are claim-specific and not predefined, how can we automatically construct diverse high-quality strategies? Second, given the limited capacity of efficient small models, how can we internalize this adaptive planning capability while preventing the generation of unstable plans?

To address these challenges, we propose a novel \textbf{L}earning to \textbf{E}valuate and \textbf{A}daptively \textbf{P}lan (\textbf{LEAP}) framework designed to endow efficient small models with the dynamic planning capabilities. 
\textbf{First}, to address the challenge of constructing diverse strategies, we establish a dynamic strategy learning loop where a teacher model iteratively explores and refines verification trajectories based on past failures. This process populates memory with high-quality strategies that transcend the limitations of fixed workflows.
\textbf{Second}, to internalize and stabilize this capability in small models, we employ agent tuning augmented by a novel proactive correction mechanism. As shown in Figure \ref{fig:intro}, a tuned critic performs a preemptive assessment of the proposed strategy's validity before tool execution. Should the initial plan be identified as suboptimal, the reflector triggers an iterative refinement loop to synthesize an optimized strategy. This ``look before it leaps'' paradigm ensures that the actor executes only precise and validated strategies, thereby achieving robust hallucination detection even within the constraints of limited model parameters.

Our contributions are summarized as follows:
\begin{itemize}
\item We propose LEAP, a new dynamic strategy learning framework that transcends fixed verification strategies and enables small models to master diverse and adaptive strategies.
\item We propose a novel proactive correction mechanism, which a tuned critic evaluates and triggers the refinement of verification strategies before execution, enhancing the robustness of strategy execution.
\item Experiments on three datasets validate the superiority of LEAP over baselines based on the fixed strategy in hallucination detection.
\end{itemize}
\section{Related Work}
\paragraph{Hallucination Detection}
Hallucination detection aims to assess the veracity of content from LLMs, a critical step to ensure their reliability \cite{luo2024hallucinationdetectionhallucinationmitigation}. Existing methods fall into two paradigms: intrinsic self-check and tool-augmented verification. Intrinsic methods operate without external knowledge, leveraging signals like token probabilities for uncertainty estimation \cite{varshney2023stitch, yao2023llm, luo2023zeroresourcehallucinationpreventionlarge}, self-consistency \cite{manakul2023selfcheckgpt,DBLP:conf/aaai/ChengTZ025}, or internal activation patterns \cite{du2024haloscope, chen2024inside, bazarova2025hallucination}. 
Although insightful, these methods fail to spot incorrect claims made with high confidence. Tool-augmented methods address this by retrieving external evidence \cite{min2023factscore, chern2023factool, dhuliawala2023chainofverificationreduceshallucinationlarge, wei2024long, xie2024fire}. 
However, these methods all adhere to a fixed verification strategy.
They execute a uniform workflow regardless of claim complexity, making them brittle against diverse hallucination patterns. 
Thus, the core challenge of learning adaptive strategies remains unsolved.

\paragraph{Agent Tuning}
Agent tuning has emerged as a powerful paradigm for allowing smaller models to learn sophisticated behaviors by finetuning them on high quality decision trajectories \cite{zeng2023agenttuning, chen2024agentflan, lai2024llmlight}. 
Although prior work has successfully distilled strategies for general purpose reasoning and planning \cite{wei2022chain, yao2023tree, shinn2023reflexion, shi2024large, bo2024reflective}, its application to hallucination detection \cite{cheng2024small} reveals a critical limitation.
Current approaches primarily focus on mimicking static verification routines. As a result, the student model learns to follow a fixed trajectory but lacks the capability to adjust when the strategy itself is flawed. 
This highlights the need to distill a strategy that is inherently dynamic and adaptive.
\section{Method}
\subsection{Problem Formulation}
Given a claim consisting of a user query $Q$ and a response $R$ generated by the model, our ultimate goal is to predict a binary label $Y \in \{\text{Hallucination}, \text{Not Hallucination}\}$. 
Rather than directly mapping the claim $(Q, R)$ to $Y$,  our approach focuses on optimizing the evidence gathering process that informs the final decision. 
A high quality process is guided by an appropriate strategy for selecting and using information gathering tools. We propose that the final verdict is determined by the quality of this evidence gathering process. 

Given our premise that strategy is critical for effective information gathering, we reframe hallucination detection as a learning problem of dynamic strategies and formalize this process through a verification strategy $\pi_{strat}$, 
which orchestrates the verification process.
Our goal is to learn an optimal strategy $\pi_{strat}^*$ that is both effective and dynamically adaptable to the specific claim.

Strategy execution relies on the collaboration of multiple specialized agents, as hallucinations are complex and diverse, requiring them to jointly operate the detection framework. Their interactions are captured in a verification trajectory $\tau$ similar to ReAct \cite{yao2024react}. A trajectory is a sequence of states and the transitions between them, composed of interleaved thoughts, actions, and observations:
\begin{equation}
\tau = (s_0, t_1, a_1, o_1, s_1, t_2, a_2, o_2, s_2, \dots, s_N)
\end{equation}
where $s_n$ represents the state at step $n$, which includes the initial claim and the history of all prior steps $(t_i, a_i, o_i)_{i=1}^{n-1}$. 
Specifically, the trajectory components are: Thought ($t_i$), the explicit reasoning guided by the overarching strategy $\pi_{strat}$ analyzing the current state to decide the next action; Action ($a_i$), a concrete operation typically involving an external tool call; Observation ($o_i$), the output returned from the tool after executing action $a_i$.
The final verdict $Y$ is determined by the terminal state $s_N$. Since the initial strategy $\pi_{strat}$ orchestrates the interactions generating this trajectory, the final verdict is a direct result of the initial plan. 

Therefore, the core challenge shifts from learning a simple classifier to discovering an optimal verification strategy $\pi_{strat}^*$, that consistently guides the agent interaction to a correct and robust verdict.

\begin{figure*}[t] 
    \centering 
    \includegraphics[width=\textwidth]{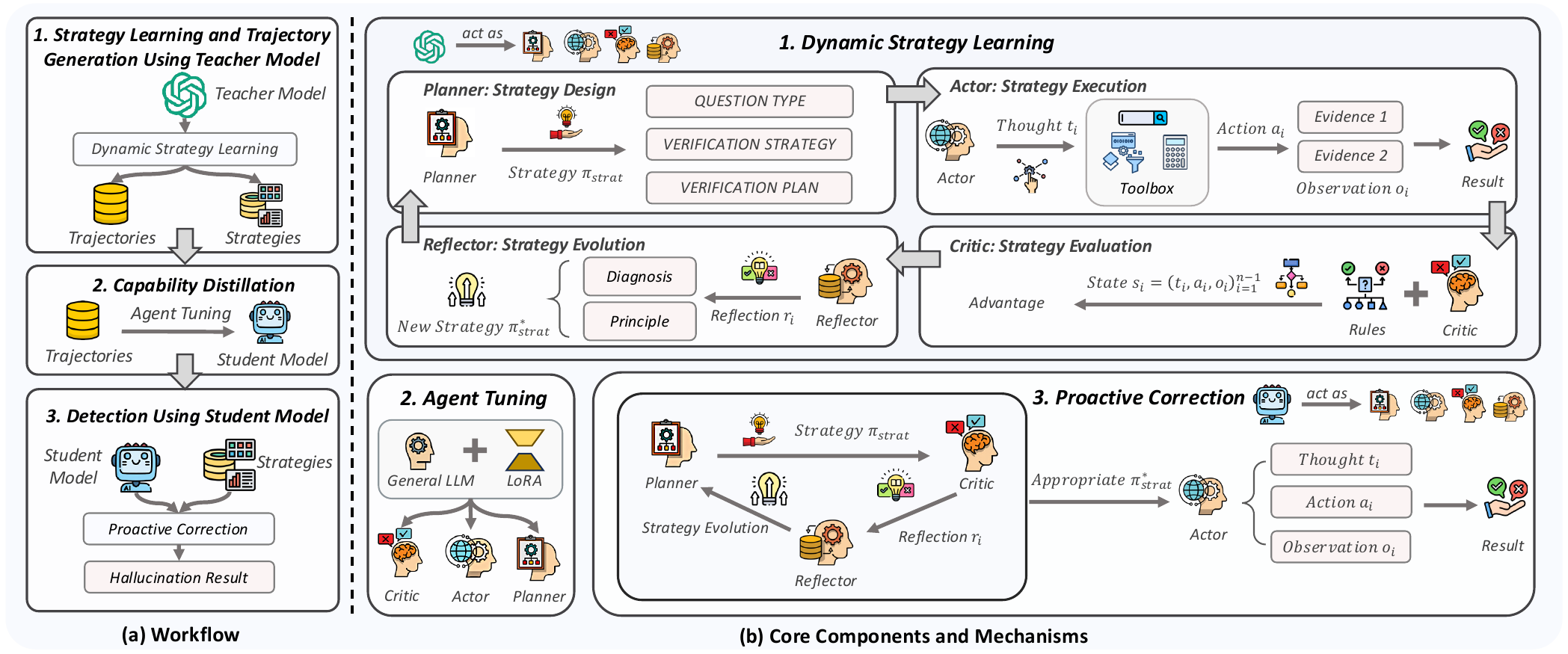} 
    \caption{The LEAP framework with its (a) workflow and (b) core components, including three main steps: 
1. Dynamic Strategy Learning and Trajectory Generation Using Teacher Model: A teacher model uses the dynamic learning loop to learn from failure and generate trajectories. 
2. Capability Distillation via Agent Tuning: The trajectories are distilled into an efficient student model. 
3. Detection with Proactive Correction Using Student Model: The student model adaptively refines its plan before execution to ensure appropriate strategies.}
    \label{fig:framework}
\end{figure*}

\subsection{Overview}
To overcome the insufficient adaptability of the fixed verification strategy, we propose LEAP, 
a framework that shifts the paradigm from fixed execution to dynamic strategy learning.
As illustrated in Fig.~\ref{fig:framework}, LEAP consists of three stages. 
\textbf{First}, to gather diverse strategies, we employ dynamic strategy learning, using \textit{a teacher model} within a dynamic learning loop that systematically learns from execution failures to continually generate superior strategies. 
\textbf{Second}, to transfer dynamic learning capabilities into \textit{an efficient small student model}, we utilize agent tuning by training on specific expert trajectories. 
\textbf{Finally}, to ensure that the strategy is adaptive in dynamical execution environments, we introduce a proactive correction mechanism, which can preemptively evaluate and optimize its own verification strategies for each specific claim before execution, thus ensuring robust performance.

\subsection{Dynamic Strategy Learning}
The first phase is dynamic strategy learning designed to generate the diverse verification strategies.
We achieve this through a closed loop where four agents collaborate to systematically learn from execution failures: the planner designs a strategy, the actor executes it to produce a trajectory, the critic evaluates the outcome, and for any failures, the reflector generates corrective feedback that is fed back to the planner.
This iterative process not only drives the continuous evolution of strategies, but also yields the high quality trajectories required for the subsequent agent tuning phase.

\paragraph{Planner: Strategy Design}
The planner is an agent responsible for creating a high-level customized verification strategy $\pi_{strat}$ for an input claim, using past experiences to go beyond fixed strategies. It generates the strategy as follows:
\begin{equation} \label{eq:planner_main}
\pi_{strat} = \text{Planner}(s_0, \mathcal{P}_p, \mathcal{R}_{\text{retrieved}}),
\end{equation}
where $s_0$ is the initial state containing the claim, $\mathcal{P}_p$ is the prompt with task instructions, and $\mathcal{R}_{\text{retrieved}}$ is a set of relevant reflections retrieved from memory. A complete strategy $\pi_{strat}$ usually specifies the type of problem, a high-level verification strategy, and a concrete verification plan.

To inform its planning, the planner performs reflection retrieval, querying a memory $\mathcal{M}_P$ to find the $K$ most relevant past reflections:
\begin{equation} \small
    \mathcal{R}_{retrieved} = \{r \mid \text{rank}(\text{sim}(\mathbf{e}_r, \mathbf{e}_{s_0})) < K, r \in \mathcal{M}_{P}\}
\end{equation}
where $\mathbf{e}$ denotes the text embedding and $\text{sim}(\cdot,\cdot)$ represents cosine similarity, implemented with FAISS \cite{johnson2019faiss} for efficient retrieval. The retrieved reflections provide relevant experiences for the planner to synthesize a new strategy. When the reflector generates a new reflection $r_{\text{new}}$ from a failure, it is integrated into the memory: 
\begin{equation}
\mathcal{M}_{P} \leftarrow \mathcal{M}_{P} \cup \{r_{\text{new}}\}.
\end{equation}

\paragraph{Actor: Strategy Execution}
The actor executes the verification strategy $\pi_{strat}$ by generating the verification trajectory $\tau$. To do so, it performs a series of actions, primarily by invoking external tools. We equip the actor with a versatile toolbox inspired by previous work \cite{cheng2024small}, including search engine, calculator, code interpreter, etc. 
Details are available in the Table~\ref{tab:toolkit}.

At each step $n$, the action $a_n$ is determined based on the strategy $\pi_{strat}$, the current state $s_n$, and a dynamically retrieved set of relevant examples $\Psi^{n}$:
\begin{equation} \label{eq:actor_main}
a_n = \text{Actor}(s_n, \pi_{strat}, \mathcal{P}_A, \Psi^{n}),
\end{equation}
where $\Psi^{n}$ is retrieved based on similarity to the current claim from the actor's memory $\mathcal{M}_A$, which stores tuples of (claim, strategy, advantage value). 
Based on the stored advantage value $A$, it's composed of samples from both successful precedents $\Psi_{\text{pos}}^{n}$ for $A > 0$ and cautionary tales $\Psi_{\text{neg}}^{n}$ for $A \leq 0$.
Upon completing the trajectory $\tau$, the actor passes it to the critic for evaluation. The actor's memory is updated with the strategy and its evaluated advantage:
\begin{equation} \label{eq:actor_update}
\mathcal{M}_A \leftarrow \mathcal{M}_A \cup \{ (s_0, \pi_{strat}, A(\pi_{strat}, \tau)) \},
\end{equation}
where $A(\pi_{strat}, \tau)$ is the advantage value calculated by the critic.

\paragraph{Critic: Strategy Evaluation}
The critic is an evaluator agent that provides a quantitative feedback signal by calculating the advantage value $A(\pi_{strat}, \tau)$ for a completed trajectory $\tau$. Once the trajectory is finalized, the critic assigns a comprehensive strategic score $V(s_0)$ based on the predefined scoring principles illustrated in Figure \ref{fig:critic_prompt}.

To formalize this process, the critic estimates a state-value function $V(s_n)$, representing the measured utility of the trajectory from state $s_n$. The critic models this function by using its memory $\mathcal{M}_C$ to store past (state, value) pairs, allowing it to fit the value function via in-context learning:
\begin{equation} \label{eq:critic_value_estimation}
V(s_n) = \text{Critic}(s_n, \mathcal{P}_C, \mathcal{M}_C).
\end{equation}
After each trajectory, the newly computed state values are used to update the memory: 
\begin{equation}
\mathcal{M}_C \leftarrow \mathcal{M}_C \cup \{ (s_n, V(s_n)) \mid s_n \in \tau \}.
\end{equation}

With the learned value function, 
we compute the advantage by adapting the commonly used advantage function~\cite{sutton1999policy} to our task:
\begin{equation} \small
\label{eq:advantage_calculation}
A(\pi_{strat}, \tau) = R_T - V(s_0) - \lambda \cdot N_{\text{tools}},
\end{equation}
where $R_T$ denotes detection success, $V(s_0)$ denotes the strategic quality score and $\lambda \cdot N_{\text{tools}}$ penalizes redundancy to ensure a parsimonious yet accurate verification path. This advantage $A$ serves as the core signal for strategy optimization and subsequent agent distillation.

\paragraph{Reflector: Strategy Evolution}
The reflector is the engine agent for strategy evolution, operating specifically on failures. When a trajectory is assigned a negative value by the critic, the reflector generates corrective feedback:
\begin{equation} \label{eq:reflector_main}
r_{\text{new}} = \text{Reflector}(\tau_{\text{fail}}, \mathcal{P}_{R}),
\end{equation}
using a specialized prompt $\mathcal{P}_{R}$, the reflector analyzes the failed trajectory $\tau_{\text{fail}}$ and generates a structured reflection $r_{\text{new}}$, which contains diagnosis of failures, generalizable high-level principles to prevent similar errors, and a revised verification strategy. The new reflection $r_{\text{new}}$ is then added to the planner's memory $\mathcal{M}_P$, directly closing the learning loop and systematically converting failures into improved future strategies.

Through this loop, we curate a pool of 1,889 distinct strategies that cover a range of patterns, including entity validation and complex contextual synthesis. Details are in the Appendix \ref{app:strategy_diversity}.

\subsection{Agent Tuning}
The second phase is agent tuning, where we finetune efficient small models using the trajectories collected in the first phase. The trajectories provide the entire reasoning process, allowing the student model to learn how to plan, not just what the result is.
This process distills dynamic learning capabilities, resulting in a small but powerful detector capable of adaptive learning.

\paragraph{Trajectory Collection} First, we employ the LEAP framework that uses a powerful model to process verification tasks from commonly used benchmarks such as HaluEval \cite{li2023halueval} and MMLU-Pro \cite{wang2025mmlu}. 
To ensure the quality of training data, we perform a strict curation process. We retain only the trajectories, those that not only reach the correct final verdict but also exhibit high efficiency. This quality is quantified by the advantage value computed by the critic; we filter for trajectories where $A(\pi_{strat}, \tau)$ is positive. This process yields high quality trajectories $\mathcal{D}_{\text{expert}}$.

\paragraph{Trajectory Finetuning}
To mitigate the planning instability in small models, we implement functional specialization by training distinct LoRA adapters~\cite{hu2022lora} for the planner, actor, and critic. This decoupling prevents interference between capabilities, allowing the system to dynamically orchestrate these specialized models. 
For each trajectory $\tau \in \mathcal{D}_{expert}$, we format it as a multi-turn conversation sequence consisting of the initial state $s_0$ and the full history of interleaved thoughts, actions, and observations $(t_1, a_1, o_1, \dots, t_T, a_T, o_T)$. 

To finetune the planner and actor, we employ a standard instruction-tuning objective where the model takes the full history as context but computes the loss only on the thought and action tokens generated by the agent, masking the observation tokens provided by tools. The objective is defined as:
\begin{equation}\small
    \mathcal{L}_{SFT}(\theta) = -\sum_{(s_0, \tau) \in \mathcal{D}} \sum_{j \in \mathcal{I}_{agent}} \log P_{\theta}(y_j | s_0, y_{<j}),
\end{equation}
where $y$ represents the linearized sequence of all tokens in the trajectory, and $\mathcal{I}_{agent}$ denotes the indices of tokens belonging to thoughts and actions.

For the critic, we construct a specialized dataset $\mathcal{D}_{crit} = \{(s_0, \pi_{strat}, A)_i\}$, where the labels $A$ are advantage values collected from the teacher’s actual executions. By predicting these values from the strategy alone, the critic internalizes the mapping between strategic logic and eventual success. This capability provides the predictive foundation for the proactive correction mechanism to stabilize planning during inference.

\subsection{Proactive Correction}
The final phase is the proactive correction mechanism designed to ensure that the tuned small model adaptively optimizes its strategy for each specific claim, thereby mitigating the planning instability.

Given a claim $s_0$, the finetuned planner first generates an initial strategy $\pi_{strat}$. Critically, rather than proceeding to immediate execution, LEAP intercepts this plan for a preemptive evaluation. The finetuned critic assesses the strategy's quality by predicting an estimated advantage score:
\begin{equation} \label{eq:preemptive_eval}\small
\hat{A}(\pi_{strat}) = \text{Critic}_{\text{tuned}}(s_0, \pi_{strat}, \mathcal{P}_C, \mathcal{M}_C),
\end{equation}
where $\hat{A}(\pi_{strat})$ predicts the likely success and efficiency of the proposed strategy. This is different from conventional post-hoc reflection by identifying potential reasoning flaws before tool calls.

The predicted advantage is then compared to a confidence threshold $\theta_{\text{corr}}$. If the strategy is deemed sufficiently robust (i.e. $\hat{A}(\pi_{strat}) \ge \theta_{\text{corr}}$), it is approved for the actor. Otherwise, the proactive correction loop is triggered. 
The reflector diagnoses the strategy's weaknesses and generates corrective feedback $r_{\text{corr}}$, which guides the planner to synthesize a revised and superior strategy $\pi'_{strat}$.
This iterative refinement ensures that the finetuned actor executes only validated and precise strategies. Once a strategy is approved, the actor takes over. It executes the strategy, generating the thought-action trace $\tau'$ by interacting with the necessary tools to reach the final detection verdict.
\section{Experiments}

\subsection{Experimental Setup}
\paragraph{Datasets and Metrics}
To comprehensively evaluate LEAP, we use three challenging benchmarks with strictly non-overlapping splits: HaluEval~\cite{li2023halueval} and MMLU-Pro~\cite{wang2025mmlu} as in-domain datasets and XTRUST~\cite{li2024xtrust} as out-of-domain datasets to evaluate robustness. Following prior work~\cite{xie2024fire, cheng2024small}, we strictly curate the test sets with hallucination ratios ranging from 44\% to 55\% to prevent majority-class bias, using accuracy and F1 score as evaluation metrics. Details are in the Appendix \ref{sec:experimental_details}.

\begin{table*}[t]
\centering
\small
\renewcommand{\arraystretch}{0.95} 
\begin{tabular}{clcccccccc}
\toprule
\multirow{3}{*}{\textbf{Models}} & \multirow{3}{*}{\textbf{Methods}} & \multicolumn{4}{c}{\textbf{In-Domain}} & \multicolumn{2}{c}{\textbf{Out-of-Domain}}  & \multicolumn{2}{c}{\multirow{2}{*}{\textbf{Average}}} \\
\cmidrule(lr){3-6} \cmidrule(lr){7-8}
& & \multicolumn{2}{c}{\textbf{HaluEval}} & \multicolumn{2}{c}{\textbf{MMLU-Pro}} & \multicolumn{2}{c}{\textbf{XTRUST}} & \multicolumn{2}{c}{} \\
\cmidrule(lr){3-10}
 & & \textbf{Acc} & \textbf{F1} & \textbf{Acc} & \textbf{F1}  & \textbf{Acc} & \textbf{F1} & \textbf{Acc} & \textbf{F1} \\
\toprule
\multirow{10}{*}{Qwen2.5-7B} 
& Perplexity & 56.33 & 57.05 & 57.33 & 68.00 & 47.50 & 62.63 & 54.50 & 62.55 \\
& LN-entropy & 54.67 & 66.00 & 56.67 & 67.50 & 48.00 & 62.32 & 53.75 & 65.64 \\
& Semantic Entropy & 50.67 & 66.67 & 55.67 & 66.67 & 47.00 & 61.31 & 51.63 & 65.33 \\
& Self-Check(0) & 51.20 & 63.40 & 56.66 & \underline{69.54} & 46.07 & 59.92 & 52.00 & 64.97 \\
& Self-Check(3) & 55.94 & 62.30 & 59.56 & 64.52 & 52.52 & 60.24 & 56.70 & 62.74 \\
\cmidrule[0.02em](lr){2-10}
& FACTOOL & 59.00 & 67.20 & 59.60 & 69.23 & 47.42 & 56.78 & 56.38 & 65.53 \\
& SAFE & 57.60 & 66.10 & 55.81 & 68.28 & 44.25 & 54.03 & 53.73 & 64.25 \\
& FIRE & 65.67 & 68.31 & \underline{61.05} & 69.23 & 53.61 & 57.94 & 60.97 & \underline{66.13} \\
\cmidrule[0.02em](lr){2-10}
& HaluAgent & \underline{70.55} & \underline{71.33} & 54.68 & 55.00 & \underline{61.93} & \underline{64.11} & \underline{62.58} & 63.62 \\
& LEAP & \textbf{74.19} & \textbf{75.00} & \textbf{69.81} & \textbf{75.31} & \textbf{64.00} & \textbf{66.36} & \textbf{69.89} & \textbf{72.88} \\
\midrule
\multirow{10}{*}{Llama3.1-8B} 
& Perplexity & 50.00 & 66.67 & 55.00 & \underline{70.97} & 44.00 & 61.11 & 50.38 & 66.89 \\
& LN-entropy & 57.34 & 66.14 & 58.34 & 68.51 & 46.00 & 61.70 & 54.88 & 65.92 \\
& Semantic Entropy & 50.67 & 66.96 & 55.33 & 65.59 & 43.00 & 58.99 & 50.50 & 64.45 \\
& Self-Check(0) & 51.37 & 67.43 & 54.88 & 70.74 & 44.16 & 61.27 & 50.89 & 67.23 \\
& Self-Check(3) & 49.82 & 64.47 & 55.25 & 70.67 & 46.24 & 61.83 & 51.05 & 66.37 \\
\cmidrule[0.02em](lr){2-10}
& FACTOOL & 50.71 & 66.83 & 55.33 & 68.84 & 48.18 & 64.68 & 52.16 & 67.24 \\
& SAFE & 60.64 & 70.56 & 57.96 & 70.49 & 60.58 & \textbf{66.25} & 59.64 & \textbf{69.75} \\
& FIRE & 65.96 & 70.73 & \underline{58.85} & 69.51 & 58.09 & \underline{65.87} & 61.72 & 69.26 \\
\cmidrule[0.02em](lr){2-10}
& HaluAgent & \underline{69.36} & \textbf{71.92} & 54.41 & 54.05  & \underline{63.30} & 59.65 & \underline{63.36} & 62.92 \\
& LEAP & \textbf{70.00} & \underline{71.88} & \textbf{64.23} & \textbf{71.18} & \textbf{64.50} & 63.59 & \textbf{66.54} & \underline{69.71} \\
\midrule
\multirow{10}{*}{Mistral-8B} 
& Perplexity & 50.00 & 66.67 & 56.67 & 67.17 & 44.00 & 61.11 & 51.00 & 65.47 \\
& LN-entropy & 56.67 & 69.48 & 58.67 & 68.04 & 44.50 & \underline{61.32} & 54.38 & 66.90 \\
& Semantic Entropy & 51.33 & 66.67 & 54.67 & 69.64 & 41.00 & 57.55 & 50.00 & 65.50 \\
& Self-Check(0) & 51.67 & 61.74 & 60.14 & 70.37 & 44.95 & 54.77 & 53.02 & 63.33 \\
& Self-Check(3) & 55.70 & 65.63 & 59.12 & 70.42 & 47.74 & 61.19 & 54.98 & 66.35 \\
\cmidrule[0.02em](lr){2-10}
& FACTOOL & 62.02 & 68.95 & \underline{62.26} & \underline{71.43} & \underline{48.37} & 55.87 & 59.15 & 67.27 \\
& SAFE & 55.05 & 67.83 & 60.61 & \textbf{72.19} & 45.16 & 59.72 & 54.96 & \underline{67.75} \\
& FIRE & 53.00 & 66.33 & 58.23 & 71.11 & \underline{48.37} & 60.70 & 53.87 & 66.95 \\
\cmidrule[0.02em](lr){2-10}
& HaluAgent & \underline{73.49} & \underline{72.85} & 54.29 & 70.37 & 46.43 & 59.46 & \underline{62.75} & 66.55 \\
& LEAP & \textbf{74.00} & \textbf{74.17} & \textbf{63.21} & 71.15  & \textbf{57.50} & \textbf{61.54} & \textbf{65.90} & \textbf{69.77} \\
\bottomrule
\end{tabular}
\caption{Main results on three hallucination detection datasets. We compare LEAP against state-of-the-art baselines across three open-source LLMs. \textbf{Bold} denotes the best performance, and \underline{underline} indicates the second-best.}
\label{tab:main}
\end{table*}

\paragraph{Baselines}
We compare against state-of-the-art baselines, including (1) intrinsic methods (Perplexity~\cite{ren2023ppl}, LN-entropy~\cite{Malinin2021LN-entropy}, Semantic Entropy~\cite{Kuhn2023SemanticUncertainty}, Self-CheckGPT \cite{manakul2023selfcheckgpt}); (2) tool-augmented prompt-based methods (Factool \cite{chern2023factool}, SAFE \cite{wei2024long}, FIRE \cite{xie2024fire}) and (3) finetuned methods (HaluAgent \cite{cheng2024small}). 
Details are in the Appendix~\ref{sec:baselines}.

\paragraph{Implementation Details}
We employ GPT-4o mini as the teacher model to generate diverse trajectories with decoding temperature 1.0 and top-p 1.0, selected for its proven high capability in fact checking \cite{xie2024fire}.
We distill three open source models: Qwen2.5-7B-Instruct \cite{qwen2025qwen25technicalreport}, Llama3.1-8B-Instruct \cite{grattafiori2024llama}, and Mistral-8B-Instruct \cite{jiang2024mixtral}. The student models are finetuned on the trajectories using LoRA, with a rank of 8, $\alpha$ of 32, a learning rate of 1e-4 and hyperparameters $\lambda = 0.1$, $K = 2$, $\theta_{\text{corr}} = 0$. For fair comparison, the temperature for all models is set to 0.0 during evaluation to ensure deterministic outputs.

\subsection{Main Results and Analysis}
Table \ref{tab:main} demonstrates the consistent superiority of LEAP across all models and datasets.
On Qwen2.5-7B, our method achieves an accuracy of 69.89\%, surpassing the best baseline by a margin of 7.31\%. This performance highlights three critical insights into effective hallucination detection.
\paragraph{The necessity of external tools.} The substantial performance gap between LEAP and intrinsic methods confirms that relying solely on internal signals is insufficient. Intrinsic approaches are bounded by the model's parametric knowledge and fail when the model generates incorrect information with high confidence, necessitating external evidence for reliable verification.
\paragraph{Advantage of dynamic planning over fixed strategies.} Comparisons with tool-augmented prompt-based baselines reveal that tool access alone is inadequate without adaptive strategies. Fixed pipelines like Factool and SAFE mechanically execute predefined workflows, which prove brittle against relational hallucinations where errors stem from flawed logical connections rather than simple factual inaccuracies. LEAP overcomes this fixity by employing dynamic strategy learning, enabling the model to tailor its verification plan to the specific logical structure of each claim.
\paragraph{Distilling planning capabilities versus mimicking execution.} Crucially, LEAP significantly outperforms HaluAgent, the strongest finetuned baseline. While HaluAgent learns to mimic a fixed execution procedure, it inherits the limitations of the fixed pipeline. In contrast, LEAP distills dynamic planning and proactive correction capabilities from the teacher model. By training on trajectories that involve strategy evaluation and refinement, our student model learns to assess and optimize its own verification logic before execution, rather than merely following a fixed script.

\begin{table}[t]
\centering
\setlength{\tabcolsep}{1mm}
\small
\resizebox{0.48\textwidth}{!}{
    \begin{tabular}{lcccccc}
    \toprule
    \multirow{2}{*}{\textbf{Method}} & \multicolumn{2}{c}{\textbf{HaluEval}} & \multicolumn{2}{c}{\textbf{MMLU-Pro}} & \multicolumn{2}{c}{\textbf{Xtrust}}\\
    \cmidrule(l){2-3} \cmidrule(l){4-5} \cmidrule(l){6-7}
    & Acc & F1 & Acc & F1 & Acc & F1 \\
    \midrule
    \multirow{1}{*}{\textit{LEAP}} & \textbf{74.19} & \textbf{75.00} & \textbf{69.81} & \textbf{75.31} & \textbf{64.00} & \textbf{66.36} \\
    \midrule
    \multirow{1}{*}{\textit{w/o Correction}} & 71.33 & 73.46 & 66.20 & 71.55 & 63.50 & 65.07 \\
    \multirow{1}{*}{\textit{w/o Dynamic Strategy}} & 70.55 & 71.33 & 54.68 & 55.00 & 61.93 & 64.11 \\
    \multirow{1}{*}{\textit{w/o Reflection Retrieval}} & 70.00 & 71.52 & 55.59 & 58.36 & 61.00 & 65.18 \\
    \multirow{1}{*}{\textit{w/o Memory Retrieval}} & 65.33 & 67.09 & 58.19 & 61.04 & 61.81 & 65.45 \\
    \multirow{1}{*}{\textit{w/o Tools}} & 59.00 & 49.80 & 54.33 & 45.42 & 53.00 & 50.53 \\
    \bottomrule
    \end{tabular}
}
\caption{The ablation results on Qwen2.5-7B. The best results are in \textbf{bold}. 
}
\label{tab:ablation}
\end{table}

\subsection{Ablation Study}
To dissect component contributions, we conduct an ablation study using Qwen2.5-7B. As detailed in Table~\ref{tab:ablation}, we examine five variants: \textit{w/o Correction}, which disables inference-time refinement; \textit{w/o Dynamic Strategy}, which replaces the adaptive planner with a fixed pipeline; \textit{w/o Reflection Retrieval} and \textit{w/o Memory Retrieval}, which exclude past insights and execution trajectories, respectively; and \textit{w/o Tools}, removing all external tools.

The results confirm the integral role of each component. Specifically, removing memory retrieval leads to a marked decline on HaluEval, validating that accessing concrete execution examples is critical for precise tool usage in fact-centric tasks. In contrast, the exclusion of reflection retrieval hampers performance on the reasoning-heavy MMLU-Pro. This performance gap highlights that abstract insights are pivotal for guiding the planner through complex adaptation, effectively bridging the divide between fixed execution and adaptive planning. Furthermore, disabling proactive correction induces consistent degradation, verifying the value of pre-execution refinement in mitigating potential errors. Finally, substituting the dynamic strategy with a fixed pipeline results in the most severe drop of over 20\% F1 on MMLU-Pro, which demonstrates that an adaptive strategy is fundamental to overcoming the limitations of fixed paradigms.

\subsection{Analysis on Trajectory Distillation}
To assess the efficacy of trajectory distillation, we compare the performance of the Qwen2.5-7B student directly with its GPT-4o mini teacher. As shown in Table \ref{tab:4o}, the student not only achieves parity but surpasses the teacher, attaining 74.19\% accuracy on HaluEval and 69.81\% on MMLU-Pro. This performance indicates that the distillation process effectively transfers the decision-making logic and dynamic planning strategies, rather than merely mimicking outputs. Consequently, LEAP proves capable of compressing complex reasoning capabilities into a small model, enabling efficient deployment without compromising reasoning depth.

\begin{table}[t]
\centering
\setlength{\tabcolsep}{1mm}
\small
    \begin{tabular}{lcccccc}
    \toprule
    \multirow{2}{*}{\textbf{Models}} & \multicolumn{2}{c}{\textbf{HaluEval}} & \multicolumn{2}{c}{\textbf{MMLU-Pro}} & \multicolumn{2}{c}{\textbf{Xtrust}}\\

    \cmidrule(l){2-3} \cmidrule(l){4-5} \cmidrule(l){6-7}
    & Acc & F1 & Acc & F1 & Acc & F1 \\
    \midrule
    \multirow{1}{*}{{LEAP}} & 74.19 & 75.00 & 69.81 & 75.31 & 64.00 & 66.36 \\
    \midrule
    \multirow{1}{*}{{GPT 4o mini}} & 73.67 & 75.69 & 69.31 & 76.32 & 64.50 & 68.16 \\
    \bottomrule
    \end{tabular}
\caption{Comparison between Qwen2.5-7B student model and its GPT-4o mini teacher.}
\label{tab:4o}
\end{table}

\subsection{Analysis on Cross-Model Generalization}
\label{sec:cross_model}

\begin{figure}[t]
    \centering
    \includegraphics[width=\linewidth]{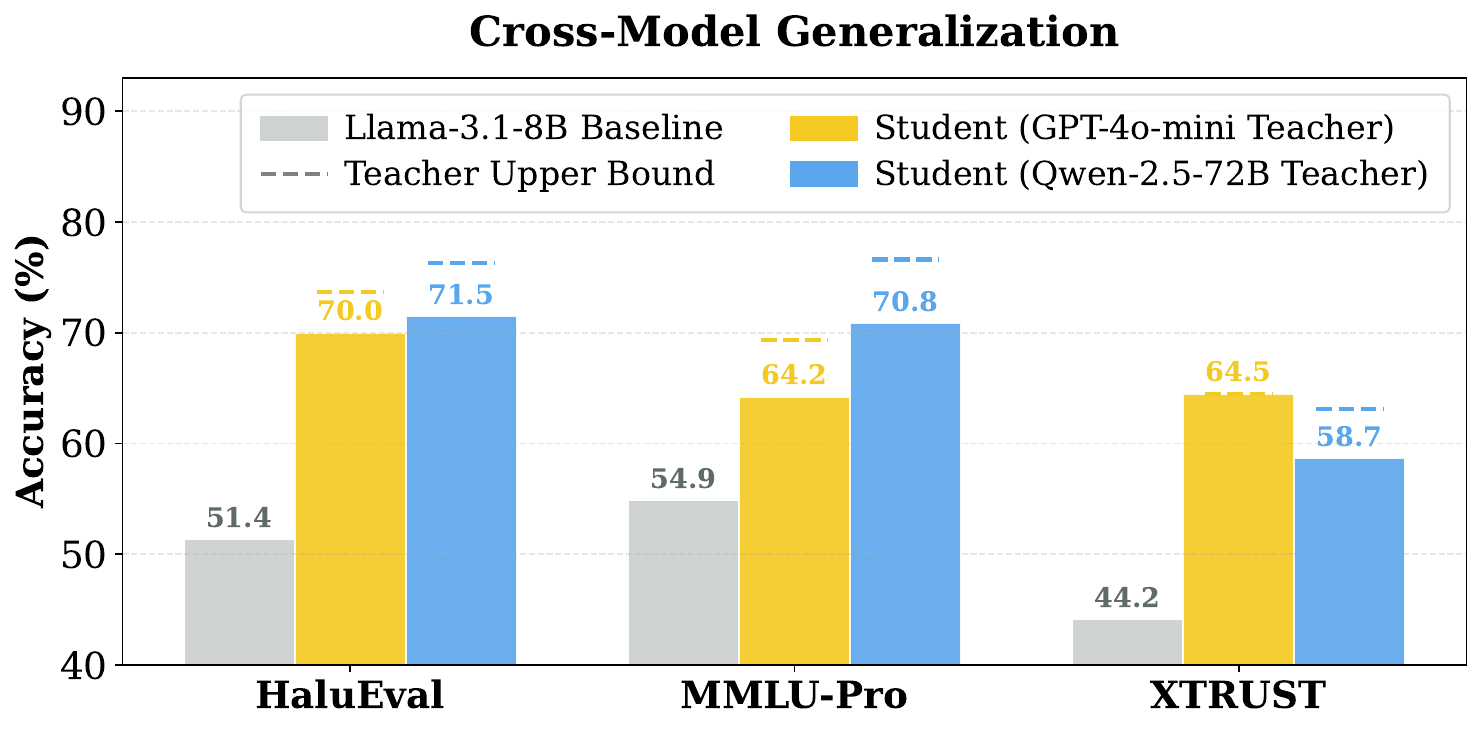}
    \caption{Cross-model generalization performance of the Llama3.1-8B student model.}
    \label{fig:cross_model}
\end{figure}

To assess architectural robustness, we use heterogeneous teacher-student pairs: Qwen2.5-72B as teacher and Llama3.1-8B as student. Figure~\ref{fig:cross_model} shows the student model achieves substantial gains over the vanilla baseline, with accuracy improvements of 20.1\% on HaluEval and 15.9\% on MMLU-Pro. Remarkably, as indicated by the teacher upper bound, the student approaches the performance of teacher model. These results validate LEAP's cross-model transferability, confirming that smaller models can effectively inherit complex dynamic planning capabilities from stronger teachers despite architectural discrepancies.

\subsection{Efficiency Analysis}
\label{sec:efficiency}

To assess the practical deployability of LEAP, we analyze its inference latency compared to HaluAgent. As shown in Table~\ref{tab:latency_tradeoff_exp}, LEAP achieves a better balance between effectiveness and efficiency. LEAP surpasses the strongest baseline by 7.31\% on Qwen2.5-7B. While LEAP increases the average latency to 18.45s from 12.32s, this overhead is intrinsic to the proactive correction mechanism where the planner and critic collaborate to optimize strategies. In high-stakes domains where reliability is paramount, this computational investment is well-justified by the significant reduction in detection failures.



\begin{table}[t]
    \centering
    \small
    \renewcommand{\arraystretch}{1.1}
    \setlength{\tabcolsep}{2.5pt} 
    \begin{tabular}{lcccc}
        \toprule
        \multirow{2}{*}{\textbf{Method}} & \multicolumn{3}{c}{\textbf{Latency (s)}} & \multirow{2}{*}{\textbf{Acc. (\%)}}  \\
        \cmidrule(lr){2-4}
         & \textbf{Min} & \textbf{Max} & \textbf{Avg} & \\
        \midrule
        HaluAgent & 8.17 & 22.21 & 12.32 & 62.58 \\
        LEAP & 10.23 & 29.10 & 18.45 & 69.89 \\
        \bottomrule
    \end{tabular}
    \caption{Inference latency and accuracy comparison on Qwen2.5-7B across three benchmarks.}
    \label{tab:latency_tradeoff_exp}
\end{table}

\begin{table}[t]
    \centering
    \small
    \setlength{\tabcolsep}{2.5pt} 
    \resizebox{\columnwidth}{!}{%
    \begin{tabular}{llcc}
        \toprule
        \textbf{Dataset} & \textbf{Method} & \textbf{Hallucinated Acc.} & \textbf{Faithful Acc.} \\
        \midrule
        \multirow{2}{*}{HaluEval} & HaluAgent & 73.29\% & 67.81\% \\
                                  & LEAP & 78.83\% & 69.72\% \\
        \midrule
        \multirow{2}{*}{MMLU-Pro} & HaluAgent & 50.99\% & 59.06\% \\
                                  & LEAP & 85.92\% & 51.22\% \\
        \midrule
        \multirow{2}{*}{XTRUST}   & HaluAgent & 76.14\% & 50.46\% \\
                                  & LEAP & 80.68\% & 50.89\% \\
        \bottomrule
    \end{tabular}%
    }
    \caption{Class-wise performance on Qwen2.5-7B.}
    \label{tab:classwise_analysis}
\end{table}

\subsection{Class-wise Performance Analysis}
\label{sec:robustness}
To ensure LEAP's gains stem from reasoning capabilities rather than a bias toward hallucination, we analyze accuracy on hallucinated versus faithful samples as in Table \ref{tab:classwise_analysis}. 
Results show LEAP excels in detecting both hallucinated and faithful content, confirming its gains are not driven by class-specific bias.
On MMLU-Pro, although LEAP's faithful accuracy declines by 7.84\% relative to HaluAgent, its hallucination detection surges by 34.93\%.
This substantial margin demonstrates proactive correction mechanism effectively identifies flaws that baselines default to as faithful, thus ensuring a robust defense against subtle errors. LEAP achieves a superior balance between hallucination and faithfulness, which significantly reduces the risk of accepting false information in high-stakes scenarios.

\section{Conclusion}
In this work, we propose LEAP, a framework that shifts tool-augmented hallucination detection from fixed execution to dynamic strategy learning. 
By integrating a dynamic strategy learning loop with a proactive correction mechanism, LEAP enables efficient small models to overcome planning instability and adaptively optimize verification strategies before execution.
Experiments on three datasets validate the superiority of LEAP, offering a scalable and reliable solution for robust hallucination detection in practical scenarios.

\newpage

\section*{Limitations}


    
While LEAP demonstrates significant improvements, it possesses inherent limitations that guide future research. First, the proactive correction mechanism introduces higher inference latency than fixed pipelines due to iterative reasoning between the critic and reflector, necessitating further optimization for ultra-low latency scenarios. Second, detection performance is bounded by external tool reliability, as noisy or outdated evidence can compromise verification quality. Finally, although GPT-4o mini serves as an effective teacher, the exploration of diverse open-source teacher models remains limited, prompting future research toward fully non-proprietary pipelines to further democratize high-quality hallucination detection.

\section*{Ethics Statement}
This work aims to enhance the reliability of LLMs by addressing the critical challenge of hallucination. We adhere to the ACL Code of Ethics and highlight the following considerations:

Enhancing Trustworthiness. Hallucinations in LLMs pose significant risks in high-stakes domains such as law and medicine. By improving the accuracy of hallucination detection through dynamic strategies, our method contributes to the development of safer and more trustworthy AI systems, mitigating the spread of misinformation.

Data and Bias. The datasets used in this study are publicly available. We acknowledge that the models may inherit biases present in these datasets or the teacher model. We have not collected any private user data, and the proposed framework is intended for research and quality assurance purposes.

\bibliography{custom}

\appendix
\newpage

\section{Experimental Details}
\label{sec:experimental_details}
\subsection{Datasets}
To comprehensively evaluate the effectiveness and generalizability of our LEAP framework, we conduct experiments on three challenging and widely recognized hallucination detection benchmarks:

\begin{itemize}
    \item \textbf{HaluEval} \cite{li2023halueval}: A general-purpose benchmark covering diverse domains and question styles (e.g., QA, dialogue) for evaluating hallucinations. It contains 35k hallucinated samples where responses were automatically generated by ChatGPT via a two-stage (sampling-then-filtering) framework. We treat HaluEval as an in-domain dataset, sampling 1,000 instances to generate expert trajectories for training and holding out 300 instances for testing.
    \item \textbf{MMLU-Pro} \cite{wang2025mmlu}: An advanced benchmark derived from MMLU, utilizing GPT-4-Turbo for option augmentation (expanding choices from 4 to 10) and Gemini-1.5-Pro for reducing false negatives. It features challenging multi-step reasoning questions across STEM, humanities, and social sciences. Similar to HaluEval, we use MMLU-Pro as an in-domain dataset, with 1,000 instances for training trajectory generation and 300 instances for testing.
    \item \textbf{XTRUST} \cite{li2024xtrust}: A benchmark focused on trustworthy evaluation across 10 languages, assessing outputs from five distinct commercial LLMs (including GPT-4, Gemini Pro, and Baichuan). It contains hard negative examples and claims requiring fine-grained grounding to external evidence, ideal for assessing detection robustness and precision. We use XTRUST as an out-of-domain dataset, evaluating on a random sample of 200 instances.
\end{itemize}

\subsection{Baselines}
\label{sec:baselines}
We compare LEAP with baselines across two distinct paradigms: intrinsic self-check and tool-augmented verification.

\noindent\textbf{Intrinsic Self-Check Methods}
These methods assess hallucinations relying on the model's internal states or outputs.
\begin{itemize}
    \item \textbf{Perplexity} \cite{ren2023ppl}: A standard metric measuring the model's confidence based on the exponentiated negative log-likelihood of the generated sequence averaged over tokens.
    \item \textbf{LN-entropy (Length-Normalized Entropy)} \cite{Malinin2021LN-entropy}: An uncertainty measure that normalizes the entropy of the predictive distribution by the sequence length, mitigating the bias where longer sequences naturally yield higher entropy.
    \item \textbf{Semantic Entropy} \cite{Kuhn2023SemanticUncertainty}: An advanced entropy-based metric that aggregates the probabilities of semantically equivalent responses (clustered via bidirectional entailment) to estimate uncertainty at the meaning level rather than the token level.
    \item \textbf{Self-CheckGPT} \cite{manakul2023selfcheckgpt}: A method that assesses hallucination by evaluating the factual consistency across multiple sampled responses to the same prompt, operating without external knowledge.
\end{itemize}

\noindent\textbf{Tool-Augmented Verification Methods}
These methods leverage external tools to verify claims.
\begin{itemize}
    \item \textbf{Factool} \cite{chern2023factool}: A framework that decomposes a response into atomic claims and verifies them using a predefined, procedural pipeline with dedicated tools like a search engine or code interpreter.
    \item \textbf{SAFE} \cite{wei2024long}: An agent-based framework that utilizes an LLM to break down a long-form response into individual facts and then iteratively issues search queries to verify the accuracy of each fact.
    \item \textbf{FIRE} \cite{xie2024fire}: A cost-effective agent that iteratively decides whether to retrieve external evidence or rely on its internal knowledge based on its confidence in the claim.
    \item \textbf{HaluAgent} \cite{cheng2024small}: A method that fine-tunes a small model on synthesized detection trajectories to act as an autonomous detector, following a verification process distilled from a teacher model.
\end{itemize}

\subsection{Implementation Details}
\subsubsection{Experimental Setup}
We employ GPT-4o mini as the teacher model to generate trajectories, selected for its proven high capability in complex fact checking tasks \cite{xie2024fire}. We then distill this capability into three leading open-source models: Qwen2.5-7B-Instruct \cite{qwen2025qwen25technicalreport}, Llama-3.1-8B-Instruct \cite{grattafiori2024llama}, and Mistral-8B-Instruct \cite{jiang2024mixtral}.

To generate a diverse set of expert trajectories, we execute our LEAP framework using the GPT-4o mini teacher with a decoding temperature of 1.0 and top-p of 1.0. After curation and filtering by the framework, this process yielded 1,075 high quality trajectories for distillation. The student models are finetuned on these trajectories using LoRA, with a rank of 8, $\alpha$ of 32, and a learning rate of 1e-4.

For a fair comparison, baselines originally designed for proprietary models were adapted to run on our open-source student models. For HaluAgent, we generated its training trajectories using the same underlying data as our method to ensure an equitable comparison. During evaluation, the temperature for all models is set to 0.0 to ensure deterministic and reproducible outputs.

\subsubsection{Experimental Environment}
For all experiments, we conduct experiments on a single Nvidia A800-80G. We use the vLLM framework \cite{kwon2023efficient} for all the LLM generation.
\subsubsection{Toolbox}
For a fair and direct comparison, we equip our method with the same versatile toolbox used in HaluAgent \cite{cheng2024small}. This includes a range of functions for both verification and system operations. A complete summary of the tools and their usage instructions is provided in Table~\ref{tab:toolkit}.

\section{Algorithms}
\label{sec:algorithms}

This section details the two core processes of the LEAP framework: (1) Dynamic Strategy Learning for offline trajectory synthesis and (2) Proactive Correction for robust inference.

\subsection{Dynamic Strategy Learning}
Algorithm~\ref{alg:dynamic_learning} outlines the closed-loop learning process of the LEAP framework. It orchestrates the collaboration between four distinct agents. The primary objective is to systematically generate a diverse set of high-quality verification strategies and their corresponding execution trajectories. By iteratively designing a strategy, executing it, evaluating the outcome, and reflecting on failures, the system continuously improves its strategic capabilities. The resulting trajectories and reflections form the high-quality data essential for the subsequent agent tuning phase.

\begin{algorithm}[t]
\caption{Dynamic Strategy Learning Loop}
\label{alg:dynamic_learning}
\begin{algorithmic}[1]
\Require States $\mathcal{S}$, agents \{Planner, Actor, Critic, Reflector\}, prompts ($\mathcal{P}_p, \mathcal{P}_A, \mathcal{P}_R, \mathcal{P}_C$), reflections $K$.
\Ensure Memories $\mathcal{M}_P, \mathcal{M}_A, \mathcal{M}_C$.
\For{each initial state $s_0 \in \mathcal{S}$}
    \State $\mathcal{R} \leftarrow \text{RetrieveReflections}(\mathcal{M}_P, s_0, K)$ 
    \State $\pi_{strat} \leftarrow \text{Planner}(s_0, \mathcal{P}_p, \mathcal{R})$ 
    \State $\tau \leftarrow \text{Actor.Execute}(\pi_{strat}, s_0, \mathcal{P}_A, \mathcal{M}_A)$ 
    \State $A \leftarrow \text{Critic.Evaluate}(\tau, \mathcal{P}_C, \mathcal{M}_C)$ 
    \State $\mathcal{M}_A \leftarrow \mathcal{M}_A \cup \{ (s_0, \pi_{strat}, A) \}$
    \State $\mathcal{M}_C \leftarrow \mathcal{M}_C \cup \{ (s_n, V(s_n)) \mid s_n \in \tau \}$
    \If{$A < 0$} 
        \State $r_{\text{new}} \leftarrow \text{Reflector}(\tau, \mathcal{P}_{R})$ 
        \State $\mathcal{M}_P \leftarrow \mathcal{M}_P \cup \{r_{\text{new}}\}$
    \EndIf
\EndFor
\end{algorithmic}
\end{algorithm}

\subsection{Proactive Correction}
Algorithm~\ref{alg:proactive_correction} details the inference-time mechanism. Instead of immediately executing a generated strategy, the system first employs a proactive correction loop. The finetuned critic preemptively assesses the initial strategy's quality. If the predicted advantage is below a confidence threshold, the reflector is triggered to provide corrective feedback, enabling the planner to revise and improve the strategy before any costly tool execution. This ensures that only high-quality strategies guide the final verification process conducted by the actor.

\begin{algorithm}[h]
\caption{Inference with Proactive Correction}
\label{alg:proactive_correction}
\begin{algorithmic}[1]
\Require Claim $s_0$, threshold $\theta_{\text{corr}}$, tuned agents.
\Ensure Final verdict $Y$.
\State $\pi_{strat} \leftarrow \text{Planner}_{\text{tuned}}(s_0)$ 
\State $\hat{A}(\pi_{strat}) \leftarrow \text{Critic}_{\text{tuned}}(s_0, \pi_{strat})$ 
\If{$\hat{A}(\pi_{strat}) < \theta_{\text{corr}}$}
    \State $r_{\text{corr}} \leftarrow \text{Reflector}_{\text{tuned}}(\pi_{strat})$ 
    \State $\pi_{strat} \leftarrow \text{Planner}_{\text{tuned}}(s_0, r_{\text{corr}})$ 
\EndIf
\State $\tau' \leftarrow \text{Actor}_{\text{tuned}}(s_0, \pi_{strat})$ 
\State $Y \leftarrow \text{GetVerdict}(\tau')$
\State \Return $Y$
\end{algorithmic}
\end{algorithm}

\section{Quantitative Analysis of Verification Strategy Diversity}
\label{app:strategy_diversity}
We conducted a comprehensive quantitative analysis on the generated dynamic strategy pool comprising 1,889 samples to verify that LEAP generates adaptive strategies rather than repeating a limited set of fixed templates.

\subsection{Semantic Projection and Clustering}
\begin{figure}[t]
    \centering
    \includegraphics[width=0.95\linewidth]{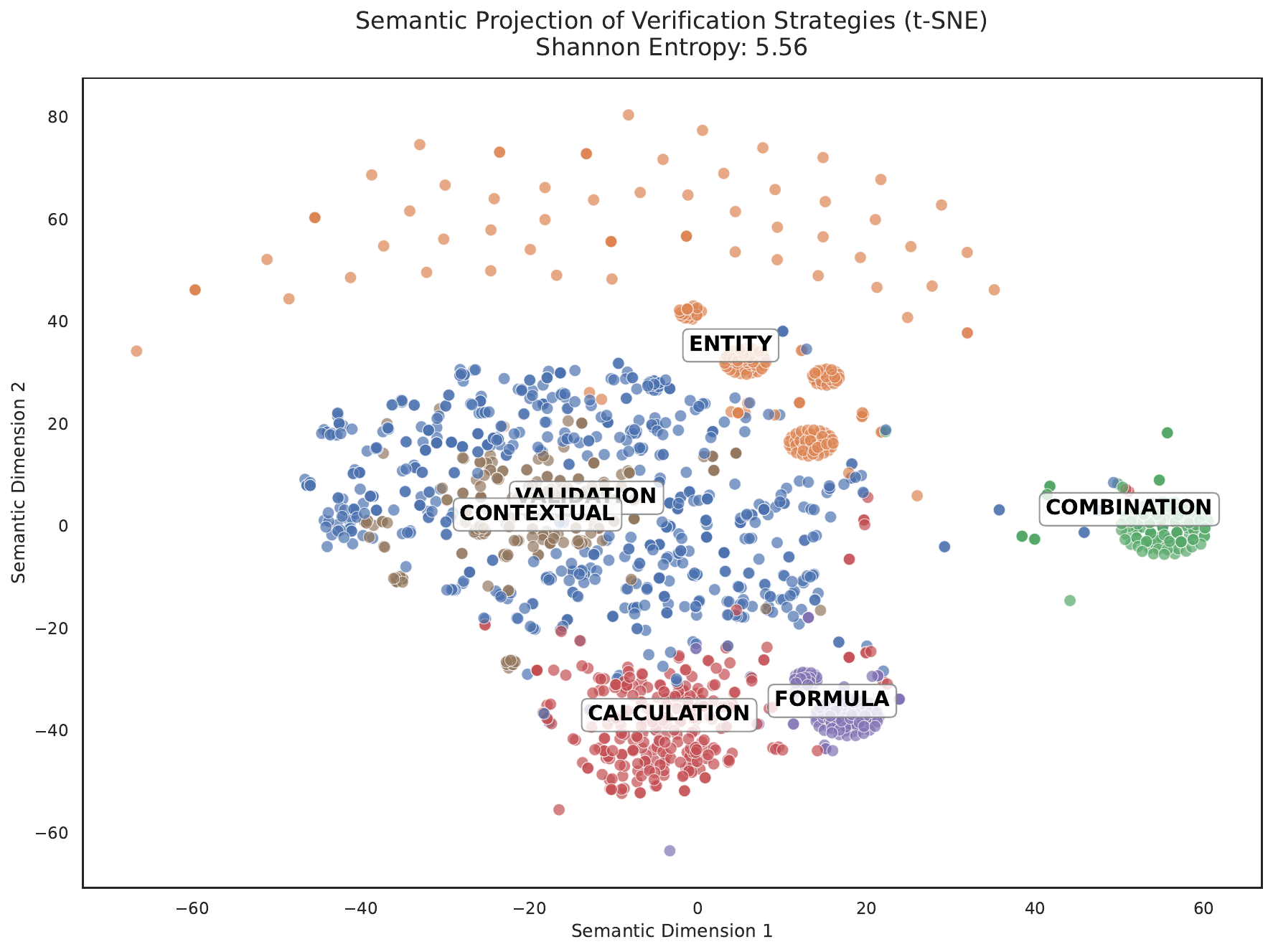} 
    \caption{Semantic projection (t-SNE) of generated verification strategies.}
    \label{fig:strategy_tsne}
\end{figure}

To visually evaluate structural diversity, we encoded all generated verification strategies into high-dimensional semantic vectors using Sentence-BERT and projected them into a 2D space using t-SNE. As shown in Figure \ref{fig:strategy_tsne}, the strategies do not collapse into a single dense region but form distinct, well-separated semantic clusters. These clusters correspond to specific reasoning capabilities required by different hallucination types:
\begin{itemize}
    \setlength\itemsep{0em}
    \item \textbf{Calculation / Formula:} Numerical verification and mathematical derivation.
    \item \textbf{Entity / Validation:} Factual entity checking and attribute verification.
    \item \textbf{Combination / Contextual:} Complex, multi-hop queries requiring information synthesis.
\end{itemize}
This semantic separation demonstrates that LEAP effectively decomposes problems into structurally distinct verification plans tailored to the specific query context.

\subsection{Quantitative Diversity Metrics}
To provide a rigorous evaluation, we calculated statistical diversity metrics as presented in Table \ref{tab:diversity_metrics}. The generated strategies exhibit a Shannon Entropy of 5.56, which is significantly higher than a system relying on fixed templates that typically yields an entropy score below 2.0. Additionally, the unique ratio of 49.13\% indicates that nearly half of the generated strategies are distinct. This confirms that the planner dynamically synthesizes verification strategies based on the specific input claim rather than mimicking static trajectories.

\begin{table}[h]
    \centering
    \small
    \renewcommand{\arraystretch}{1.3} 
    \begin{tabularx}{\linewidth}{l c X} 
        \toprule
        \textbf{Metric} & \textbf{Value} & \textbf{Interpretation} \\
        \midrule
        Total Strategies & 1,889 & Total samples collected during the dynamic learning phase. \\
        Unique Strategies & 928 & Number of distinct strategies. \\
        Diversity Ratio & 49.13\% & Ratio of unique instances. \\
        Shannon Entropy & 5.56 & Metric of distributional richness. \\
        \bottomrule
    \end{tabularx}
    \caption{Quantitative diversity metrics of the verification strategies. }
    \label{tab:diversity_metrics}
\end{table}

\begin{figure}[t]
    \centering
    \includegraphics[width=0.45\textwidth]{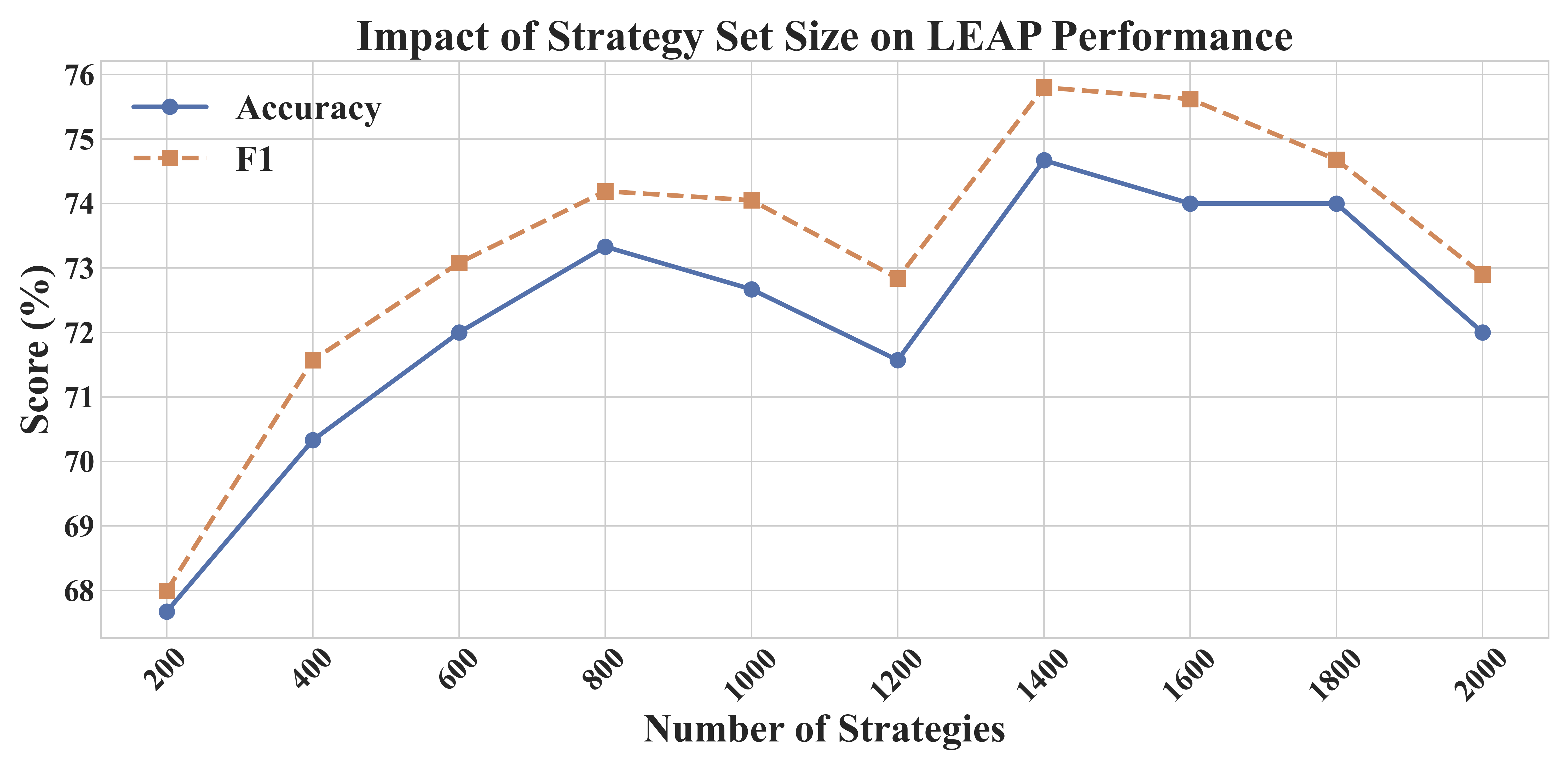}
    \caption{Performance of the Qwen2.5-7B on HaluEval as a function of strategy set size.}
    \label{fig:strategy_scale}
\end{figure}

\section{Analysis on Strategy Set Size}
To investigate the impact of the scale of strategies on performance, we varied the number of strategies available to agents. Figure~\ref{fig:strategy_scale} reveals a distinct nonmonotonic relationship. Initially, performance increases as a larger strategy pool provides greater diversity and adaptability.
Performance peaks at approximately 1,400 strategies, representing an optimal balance. Beyond this point, we observe a slight but consistent degradation. We attribute this to our retrieval mechanism. As the total number of strategies in the memory grows, the likelihood increases that similarity-based retrieval which considers only a few top-k examples may fetch less relevant or lower quality strategies into the context. This introduces noise into the planning process, potentially degrading the quality of the final generated strategy. But LEAP performance remains significantly above the baselines in Table~\ref{tab:main}, underscoring the overall robustness of our framework.

\begin{figure*}[t] 
    \centering 
    \includegraphics[width=\textwidth]{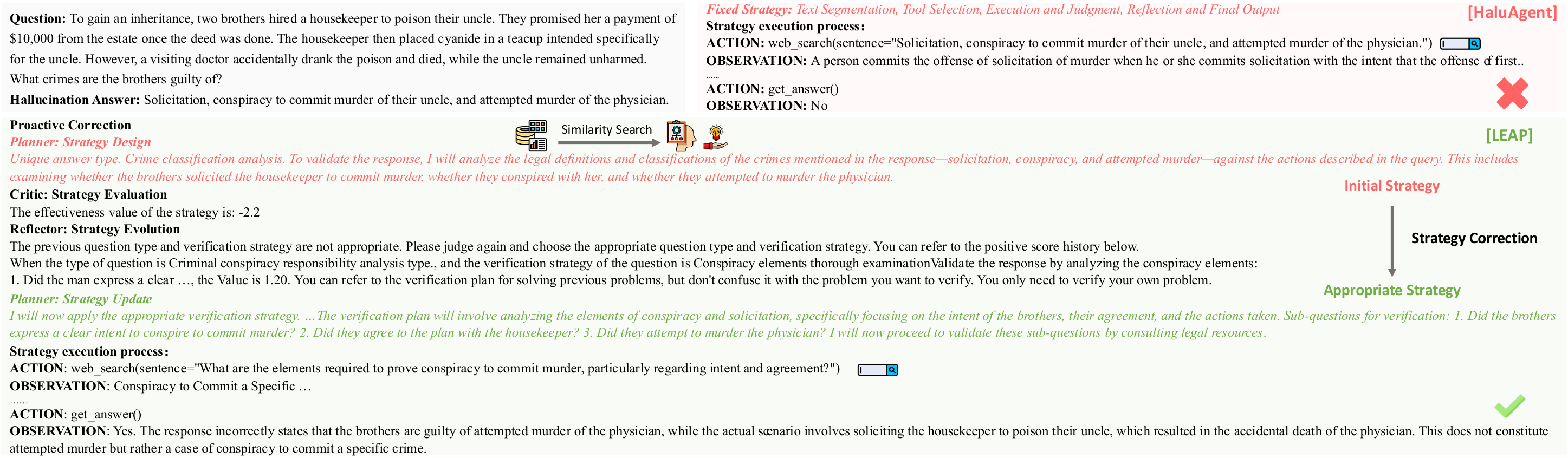}
    \caption{A case study on a complex reasoning task. HaluAgent uses its fixed strategy and LEAP uses its dynamic planning, including strategy correction and precise execution.
    }
    \label{fig:case_study}
\end{figure*}

\section{Detailed Analysis of Dataset Composition and Performance}
\label{sec:appendix_classwise_analysis}

To rigorously verify the robustness of LEAP and address potential concerns regarding dataset balance, we provide a fine-grained breakdown of dataset composition and comparative class-wise performance. We evaluate LEAP against the strongest tool-augmented baseline, HaluAgent, using both Qwen2.5-7B and Llama-3.1-8B.

\subsection{Dataset Composition}
Table~\ref{tab:dataset_composition} details the distribution of our sampled test sets. These datasets exhibit minor natural variations with hallucination ratios ranging between 44\% and 55\%. This confirms that our evaluation is conducted on balanced data, ensuring that accuracy is not skewed by majority-class dominance.

\begin{table}[h]
    \centering
    \small
    \resizebox{\columnwidth}{!}{%
    \begin{tabular}{lcccc}
        \toprule
        \textbf{Dataset} & \textbf{Total} & \textbf{Hallucinated} & \textbf{Faithful} & \textbf{Pos. Ratio} \\
        \midrule
        HaluEval & 300 & 150 & 150 & 50.0\% \\
        MMLU-Pro & 300 & 165 & 135 & 55.0\% \\
        XTRUST   & 200 & 88  & 112 & 44.0\% \\
        \bottomrule
    \end{tabular}%
    }
    \caption{Dataset composition breakdown statistics.}
    \label{tab:dataset_composition}
\end{table}

\subsection{Comparative Class-wise Performance}
To distinguish between effective strategy adaptation and baseline limitations, we analyze the accuracy on hallucinated samples versus faithful samples. The results in Table \ref{tab:comparative_classwise} reveal a strategic trade-off between LEAP and HaluAgent. While HaluAgent maintains higher accuracy on faithful samples, its verification process exhibits limited sensitivity to subtle hallucinations, as evidenced by its 46.36\% accuracy on hallucinated samples with the Llama-3.1-8B. In contrast, LEAP leverages its proactive correction mechanism to preemptively optimize verification strategies, achieving a higher 81.76\% accuracy on hallucinated samples for the same task. This substantial improvement in error detection confirms that LEAP prioritizes safety by effectively navigating intricate logical inconsistencies. Although this rigorous verification leads to a more conservative judgment on faithful content, such increased sensitivity is a deliberate design choice to ensure reliability in high-stakes scenarios where false negatives pose a critical risk.

\begin{table}[t]
    \centering
    \resizebox{\columnwidth}{!}{%
    \begin{tabular}{lllcccc}
        \toprule
        \textbf{Dataset} & \textbf{Model} & \textbf{Method} & \textbf{Acc.} & \textbf{Hallu.} & \textbf{Faith.} & \textbf{F1} \\
        \midrule
        \multirow{4}{*}{HaluEval} 
        & \multirow{2}{*}{Qwen2.5} & HaluAgent & 70.55\% & 73.29\% & 67.81\% & 71.33\% \\
        &                          & LEAP      & 74.19\% & 78.83\% & 69.72\% & 75.00\% \\
        \cmidrule{2-7}
        & \multirow{2}{*}{Llama3.1} & HaluAgent & 69.83\% & 77.03\% & 62.59\% & 71.92\% \\
        &                          & LEAP      & 70.00\% & 76.67\% & 63.33\% & 71.88\% \\
        \midrule
        \multirow{4}{*}{MMLU-Pro} 
        & \multirow{2}{*}{Qwen2.5} & HaluAgent & 54.68\% & 50.99\% & 59.06\% & 55.00\% \\
        &                          & LEAP      & 69.81\% & 85.92\% & 51.22\% & 75.31\% \\
        \cmidrule{2-7}
        & \multirow{2}{*}{Llama3.1} & HaluAgent & 56.41\% & 46.36\% & 68.85\% & 54.05\% \\
        &                          & LEAP      & 64.23\% & 81.76\% & 43.65\% & 71.18\% \\
        \midrule
        \multirow{4}{*}{XTRUST}   
        & \multirow{2}{*}{Qwen2.5} & HaluAgent & 61.93\% & 76.14\% & 50.46\% & 64.11\% \\
        &                          & LEAP      & 64.00\% & 80.68\% & 50.89\% & 66.36\% \\
        \cmidrule{2-7}
        & \multirow{2}{*}{Llama3.1} & HaluAgent & 63.30\% & 60.00\% & 66.02\% & 59.65\% \\
        &                          & LEAP      & 64.50\% & 70.45\% & 59.82\% & 63.59\% \\
        \bottomrule
    \end{tabular}%
    }
    \caption{Comparative class-wise performance. Hallu. and Faith. denote accuracy on hallucinated and faithful samples, respectively.}
    \label{tab:comparative_classwise}
\end{table}

\section{Case Study}
\label{sec:case_study}
Figure~\ref{fig:case_study} presents a complex legal case where a claim contains multiple nuanced legal concepts: solicitation, conspiracy, and attempted murder. The claim subtly misapplies the concept of ``attempted murder'' to accidental death, creating a challenging hallucination. We compare the fixed verification process of HaluAgent with the adaptive approach of LEAP.
HaluAgent hindered by its fixed strategy adopts a naive verification plan. It attempts to validate the entire claim at once, without scrutinizing its individual components.
As a result, it retrieves only general information and misses the subtle factual error regarding ``attempted murder'' thus incorrectly labeling the statement as non-hallucination.

In contrast, LEAP demonstrates a multistage adaptive process. 
The planner first designs an initial strategy to analyze the legal definitions of the three crimes mentioned. 
Critically, before execution, the proactive correction mechanism assesses this initial strategy. It identifies the strategy as suboptimal and leverages its memory from past experiences to refine it into a more precise strategy focused on verifying the core legal elements of each crime.
This optimized strategy enables a focused execution—verifying "attempted murder" as a distinct sub question—that successfully isolates and identifies the hallucination. 
This case study highlights how LEAP's ability to plan, critique, and revise its own strategy leads to a more robust detection process.

\section{Computational Cost Analysis}
\label{app:cost_analysis}

To provide a comprehensive evaluation of the economic and computational efficiency of our proposed framework, we conducted a detailed cost analysis using GPT-4o mini as a representative of efficient proprietary LLMs. Our cost calculation is based on the official pricing of OpenAI's GPT-4o mini API\footnote{\url{https://openai.com/api/pricing/}}:
\begin{itemize}
    \setlength\itemsep{0em} 
    \item Input Token Price: \$0.15 per 1M tokens.
    \item Output Token Price: \$0.60 per 1M tokens.
\end{itemize}

We performed a statistical analysis on the aggregated output logs from our experiments. The dataset comprises 1,075 valid dialogue trajectories used for training. The detailed token usage and cost breakdown are presented in Table \ref{tab:cost_breakdown}.

\begin{table}[h]
    \centering
    \small
    \renewcommand{\arraystretch}{1.2}
    \begin{tabular}{lc} 
        \toprule
        \textbf{Metric} & \textbf{Value} \\
        \midrule
        Total Dialogues & 1,075 \\
        Total Input Tokens & 3,030,168 \\
        Total Output Tokens & 417,847 \\
        \midrule
        Avg. Input Tokens per Query & 2,818.76 \\
        Avg. Output Tokens per Query & 388.69 \\
        Avg. Total Tokens per Query & 3,207.46 \\
        \midrule
        Total Cost (USD) & \$0.706 \\
        \bottomrule
    \end{tabular}
    \caption{Detailed cost analysis of the GPT-4o mini model. }
    \label{tab:cost_breakdown}
\end{table}

As shown in Table \ref{tab:cost_breakdown}, the total cost for constructing the expert dataset is approximately \$0.71. While this cost appears low, it scales linearly with request volume. In contrast, our LEAP-tuned model supports local or private cloud deployment. Once finetuned, the inference cost is decoupled from per-token commercial pricing, offering a significantly more scalable solution for high-frequency hallucination detection tasks. The LEAP framework effectively distills the reasoning capability of a paid API service into a cost-efficient model.

\section{Inference Latency Analysis}
\label{sec:latency}
To assess the practical deployability of LEAP, we analyze its inference latency compared to HaluAgent. As shown in Table~\ref{tab:latency_tradeoff}, LEAP achieves a better balance between effectiveness and efficiency. LEAP surpasses the strongest baseline by 7.31\% on Qwen2.5-7B. While LEAP increases the average latency to 18.45s from 12.32s, this overhead is intrinsic to the proactive correction mechanism where the planner and critic collaborate to optimize strategies. In high-stakes domains where reliability is paramount, this computational investment is well-justified by the significant reduction in detection failures.

\begin{table}[t]
    \centering
    \small
    \renewcommand{\arraystretch}{1.1}
    \setlength{\tabcolsep}{2.5pt} 
    \begin{tabular}{lcccc}
        \toprule
        \multirow{2}{*}{\textbf{Method}} & \multicolumn{3}{c}{\textbf{Latency (s)}} & \multirow{2}{*}{\textbf{Acc. (\%)}}  \\
        \cmidrule(lr){2-4}
         & \textbf{Min} & \textbf{Max} & \textbf{Avg} & \\
        \midrule
        HaluAgent & 8.17 & 22.21 & 12.32 & 62.58 \\
        LEAP & 10.23 & 29.10 & 18.45 & 69.89 \\
        \bottomrule
    \end{tabular}
    \caption{Inference latency and accuracy comparison on Qwen2.5-7B across three benchmarks.}
    \label{tab:latency_tradeoff}
\end{table}

\section{Instructions}
\begin{figure*}[b]
	\centering
	\includegraphics[width=0.90\textwidth]{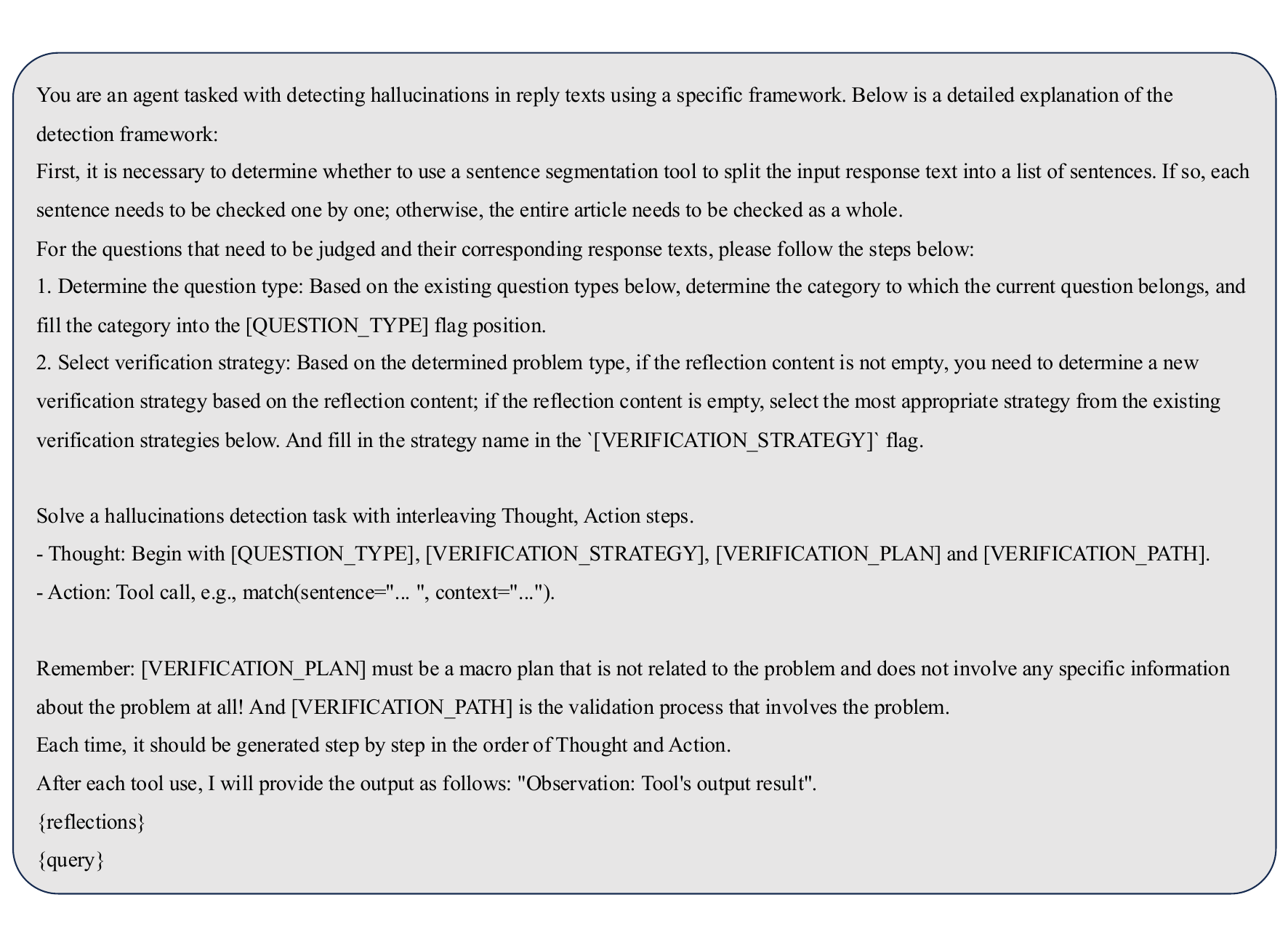}
	\caption{An example of the prompt used for the Planner agent ($\mathcal{P}_p$).}
	\label{fig:planner_prompt}
\end{figure*}

In this section, we provide the detailed instructions used to guide each agent in our framework. Each prompt is designed to elicit a specific behavior corresponding to the agent's role in the dynamic strategy learning loop.
\paragraph{Planner prompt ($\mathcal{P}_p$)}
This prompt instructs the planner on how to generate a high-level verification strategy $\pi_{strat}$ for a given claim. It guides the agent to consider the problem type, devise a general strategy, and formulate a concrete, step-by-step verification plan. The design of this prompt is crucial for generating diverse and plausible initial strategies. An example of the planner prompt is shown in Figure~\ref{fig:planner_prompt}.

\begin{figure*}[b]
	\centering
	\includegraphics[width=0.90\textwidth]{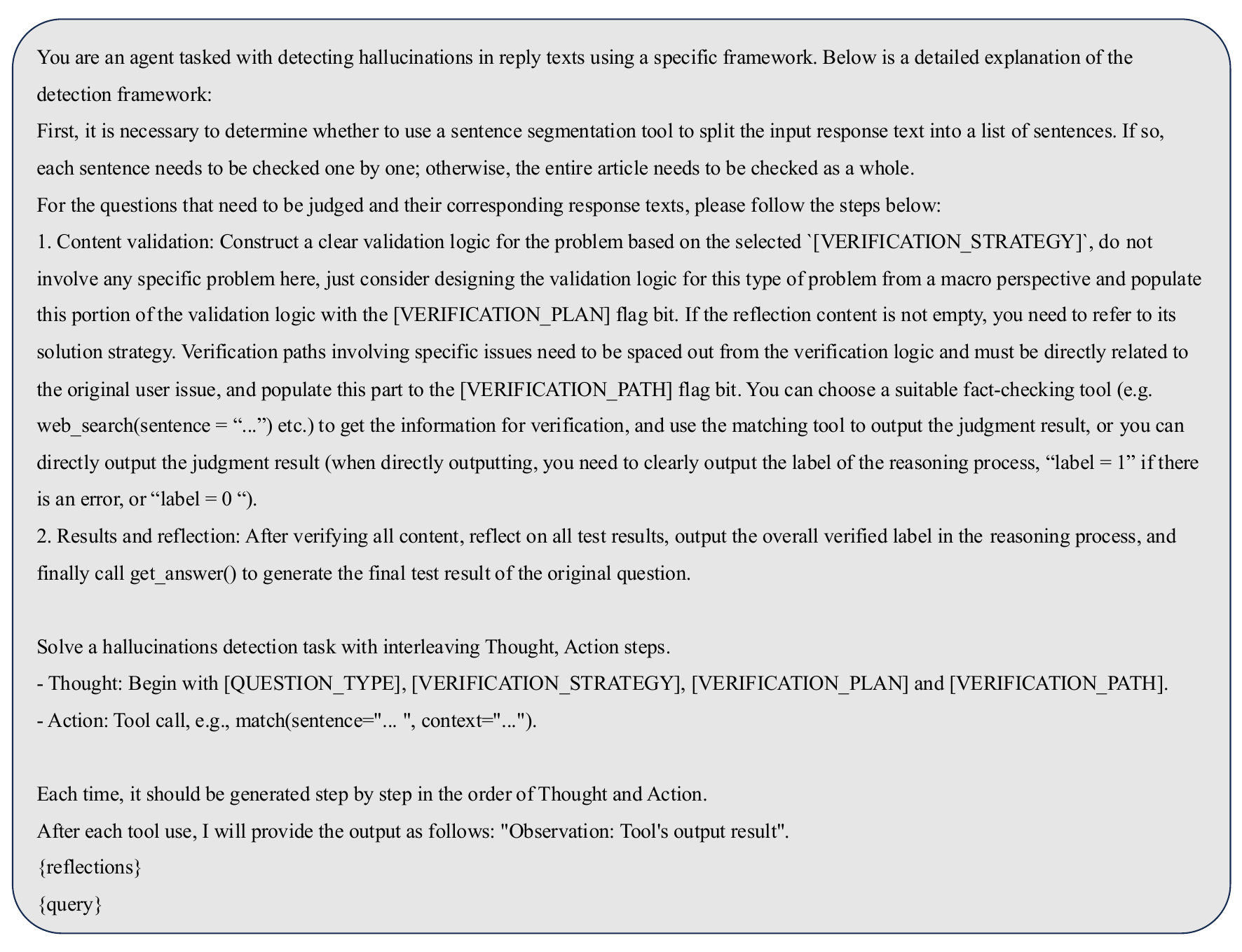}
	\caption{An example of the prompt used for the Actor agent ($\mathcal{P}_A$).}
	\label{fig:actor_prompt}
\end{figure*}

\paragraph{Actor prompt ($\mathcal{P}_A$)}
The actor prompt guides the execution agent at each step $n$ of the verification trajectory. Based on the overall strategy $\pi_{strat}$ and the current state $s_n$, this prompt asks the agent to generate the next thought and action ($t_{n+1}, a_{n+1}$). The prompt encourages the agent to make concrete tool calls to gather evidence. An example is provided in Figure~\ref{fig:actor_prompt}.

\begin{figure*}[t]
	\centering
	\includegraphics[width=0.90\textwidth]{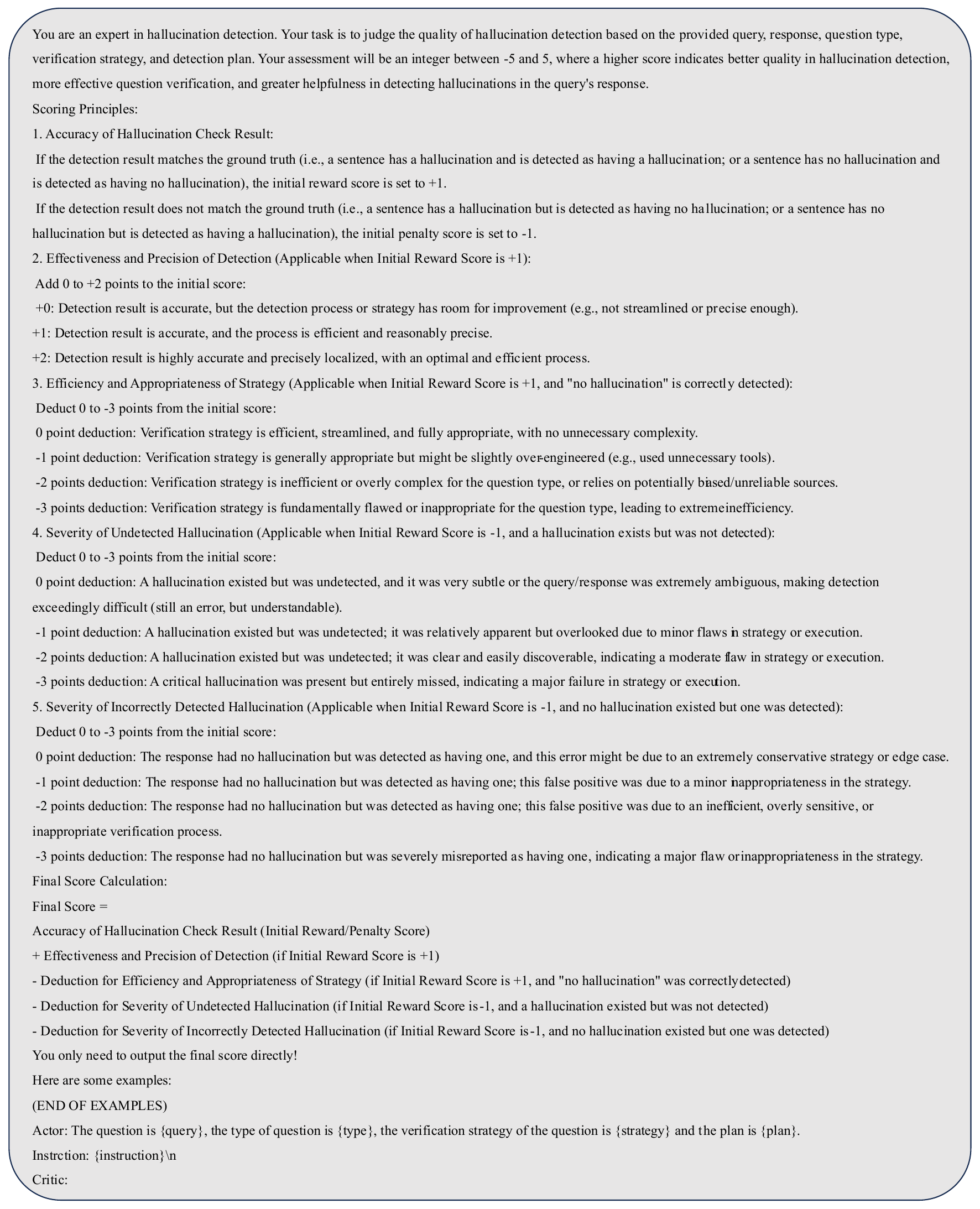}
	\caption{An example of the prompt used for the Critic.}
	\label{fig:critic_prompt}
\end{figure*}

\paragraph{Critic prompt ($\mathcal{P}_C$)}
The critic prompt is used to elicit the state-value estimation $V(s_n)$ from the critic agent. It presents the agent with a state from a trajectory and asks for a numerical evaluation of the expected future outcome. This prompt is essential for the critic to learn its value function and provide the baseline for advantage calculation. An example is shown in Figure~\ref{fig:critic_prompt}.

\begin{figure*}[t]
	\centering
	\includegraphics[width=0.90\textwidth]{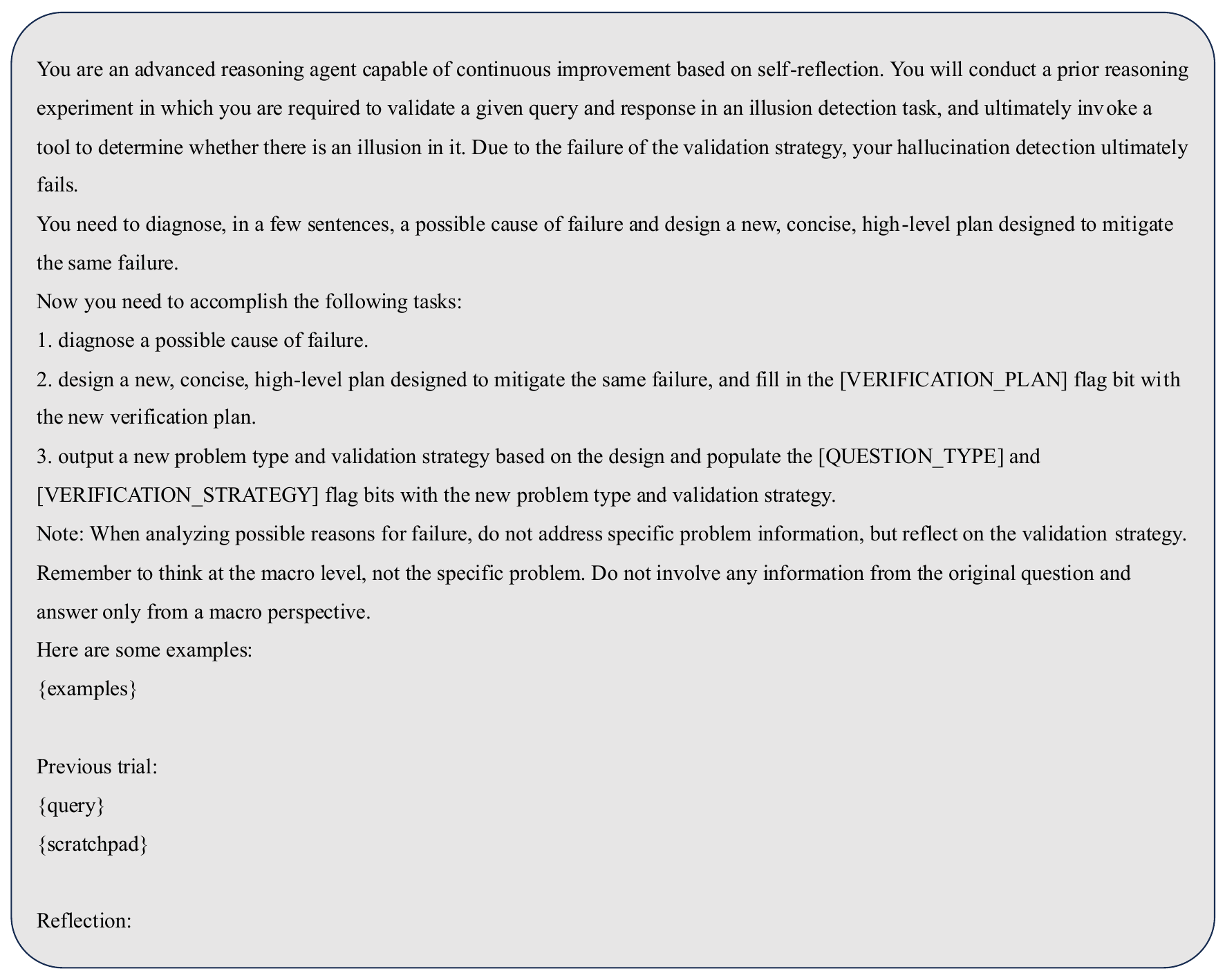}
	\caption{Description of the Reflector prompt ($\mathcal{P}_R$).}
	\label{fig:reflector_prompt}
\end{figure*}

\paragraph{Reflector prompt ($\mathcal{P}_R$)}
The reflector prompt is specifically designed to facilitate learning from failures. When a trajectory receives a negative advantage value, this prompt is used to guide the reflector. The agent analyzes the failed trajectory $\tau_{\text{fail}}$ to produce a structured reflection containing a failure diagnosis and a corrected strategy. This reflection is then stored to improve future planning. The detailed prompt is shown in Figure~\ref{fig:reflector_prompt}.

\begin{table*}[b]
\centering
\renewcommand{\arraystretch}{1.15}
\resizebox{\textwidth}{!}{
\begin{tabular}{@{}l p{12cm}@{}}
\toprule
\textbf{Tool} & \textbf{Description and Usage} \\
\midrule
\multicolumn{2}{@{}l}{\textit{\textbf{Verification Tools}}} \\
\midrule
\texttt{web\_search} & Searches the web with a query string to retrieve factual information. \\
& \textit{Usage:} \texttt{web\_search(query: str) -> str} \\
\cmidrule(l){1-2}
\texttt{calculator} & Evaluates a mathematical formula provided as a string. \\
& \textit{Usage:} \texttt{calculator(formula: str) -> float} \\
\cmidrule(l){1-2}
\texttt{code\_interpreter} & Executes a given code snippet. Returns a label indicating success or failure. \\
& \textit{Usage:} \texttt{code\_interpreter(code: str) -> bool} \\
\cmidrule(l){1-2}
\texttt{word\_count} & Counts words in a text against a specified length requirement. \\
& \textit{Usage:} \texttt{word\_count(length: int, text: str) -> (int, bool)} \\
\cmidrule(l){1-2}
\texttt{match} & Semantically matches a sentence against a provided context. \\
& \textit{Usage:} \texttt{match(sentence: str, context: str) -> bool} \\
\midrule
\multicolumn{2}{@{}l}{\textit{\textbf{System Tools}}} \\
\midrule
\texttt{split\_text} & Segments a block of text into a list of individual sentences. \\
& \textit{Usage:} \texttt{split\_text(text: str) -> list[str]} \\
\cmidrule(l){1-2}
\texttt{get\_answer} & Returns the final detection result, with optional supporting evidence. \\
& \textit{Usage:} \texttt{get\_answer() -> (str, str)} \\
\bottomrule
\end{tabular}
}
\caption{The toolbox available to our method, adapted from HaluAgent for fair comparison. It includes tools for verification and system operations.}
\label{tab:toolkit}
\end{table*}

\end{document}